\documentclass[review]{elsarticle}
\usepackage{graphicx}
\usepackage[ruled,vlined]{algorithm2e}
\usepackage{tabularx}
\usepackage{amsmath}
\usepackage{hyperref}
\usepackage{caption}
\usepackage{lineno,hyperref}
\modulolinenumbers[5]

\journal{arXiv}

\begin{document}

\begin{frontmatter}

\title{XBNet: An Extremely Boosted Neural Network}

\author{Tushar Sarkar}
\address{Analytica, Mumbai }

\begin{abstract}
Neural networks have proved to be very robust at processing unstructured data like images, text, videos, and audio. However, it has been observed that their performance is not up to the mark in tabular data; hence tree-based models are preferred in such scenarios. A popular model for tabular data is boosted trees, a highly efficacious and extensively used machine learning method, and it also provides good interpretability compared to neural networks. In this paper, we describe a novel architecture XBNet (Extremely Boosted Neural Network), which tries to combine tree-based models with neural networks to create a robust architecture trained by using a novel optimization technique, Boosted Gradient Descent for Tabular Data which increases its interpretability and performance.
\end{abstract}


\end{frontmatter}

\section{Introduction}

The need and use for data keep increasing day by day in our current world, and its impact on people's daily lives is motivating data-driven decisions in many sectors of the industry. The applications of these approaches are spam/ham detectors that are commonly used in emails, and anti-virus systems that prevent spamming of emails and protect our data and our computers from potential viruses. It is used in the marketing industry as advertising systems learn to correlate the correct advertisements with the users who are looking for items in those contexts \cite{jordan2015machine}. Fraud is prevented by top companies by using such systems that monitor all the activity and detect any vitriolic activity and alert people to prevent it promptly; anomaly detection techniques help banks to check to prevent fraudulent transactions \cite{wang2010don}. The reason behind the success of these applications is the following: the utility of statistical models that maps complex data dependencies between disparate entities.
Furthermore, the scalability of these architectures enables them to learn the relationships even in complex data sources to solve our problems and help us achieve our goals.
Even in the algorithms that are commonly used in dealing with tabular data, gradient tree boosting is an approach that outshines others in several use cases \cite{friedman2001greedy}.
Deep neural networks have shown noteworthy success
with unstructured data like images, text, videos, and audio \cite{he2015delving} \cite{devlin2018bert} \cite{lecun2015deep} \cite{mobahi2009deep}. For the above domains of problems, we try to precisely encode the information hidden in them into a vector space that forms the basis of our understanding and helps tackle the problem. Tabular data is the only data type where there is a minimum success with this approach. Though it is the most common data type in the actual world of data science, which comprises any continuous and discrete features, deep learning for tabular data remains dormant, as ensemble methods built on decision trees continue to perform better on such data \cite{chui2018notes}. Decision trees develop a piece-wise function for classifying the data and solving its objective, whereas deep neural networks develop a non-linear function for mapping the relationship between the input features and the target vectors. The activation function introduces this non-linearity in the neural network. Different types of activation functions introduce different degrees of activation of the neuron in a particular layer that enables it to grasp the relation between the input features and output vectors. Since tree-based models create a piece-wise function, it makes them more interpretable than neural networks and can also be used to understand how various features affect the dependent variable.
We propose a new architecture for tabular
data, XBNet, which attempts to combine gradient boosted trees with feed-forward neural networks to give rise to a new approach to robust architectures. XBNet inputs raw tabular data and is trained using an optimization technique Boosted Gradient Descent which is initialized with the feature importance of a gradient boosted tree, and it updates the weights of each layer in the neural network in two steps:
\begin{enumerate}
\item[(1)] Update weights by gradient descent.
\item[(2)] Update weights by using feature importance, of a gradient boosted tree in every intermediate layer \cite{hooker2018evaluating}.
\end{enumerate}

\section{Related Work}

Tree-based models and their variants like AdaBoost, Random Forest, XGBoost, etc are widely used in classification and regression problems \cite{hastie2009multi}. They repeatedly split
the input vector space and allot scores to the final node.
Tree-based models not only boost the performance in tabular data they also increase the interpretability, of the system which increases its usability in business scenarios \cite{malioutov2017learning}. It is also frequently seen that  ensemble techniques like Random Forest and XGBoost are used in most of the winning solutions in the case of tabular data \cite{pedregosa2011scikit}
\cite{breiman2001random} \cite{10.1145/2939672.2939785}. These models perform better than neural networks at several classification and regression problems when it comes to tabular data. 
Some works have advanced to amalgamate neural networks and tree-based models like decision trees to develop a better architecture with the features of both neural network and tree-based models. An approach in this direction was Neural Decision Forests that attempts to combine trees with representational learning\cite{rota2014neural}. Another unique approach is DNDT which stands for Deep Neural Decision tree. DNDTs have a unique architecture, where a particular set of weights maps to a distinct decision tree \cite{yang2018deep}. Our proposed architecture XBNet is another step in that field and it differs from the previous methods in many approaches. The optimization technique acts as an extension of any gradient-based optimization technique with the help of tree-based models \cite{ruder2016overview}. 

Recent advancement in this field is the attempt to use convolution layers for encoding the information to create a feature map which can then be fed to an extreme gradient boosted tree for inference \cite{thongsuwan2021convxgb}. A similar approach is also followed for dealing with images by using a combination of convolution layers and XGBoost to propel the research and work in the field of breast cancer detection \cite{sugiharti2021convolutional}. A convolution-based architecture coupled with a classifier for inference and prediction utilized for maximizing performance measures is also rooted in the above-discussed techniques \cite{geng2016learning}. 

In our approach, trees are trained in every layer of the architecture and their feature importance is used along with the weights determined by gradient descent for adjusting the weights of those layers respectively.

\begin{figure*}[t!]
\centerline{\includegraphics[width=12cm, height = 9cm]{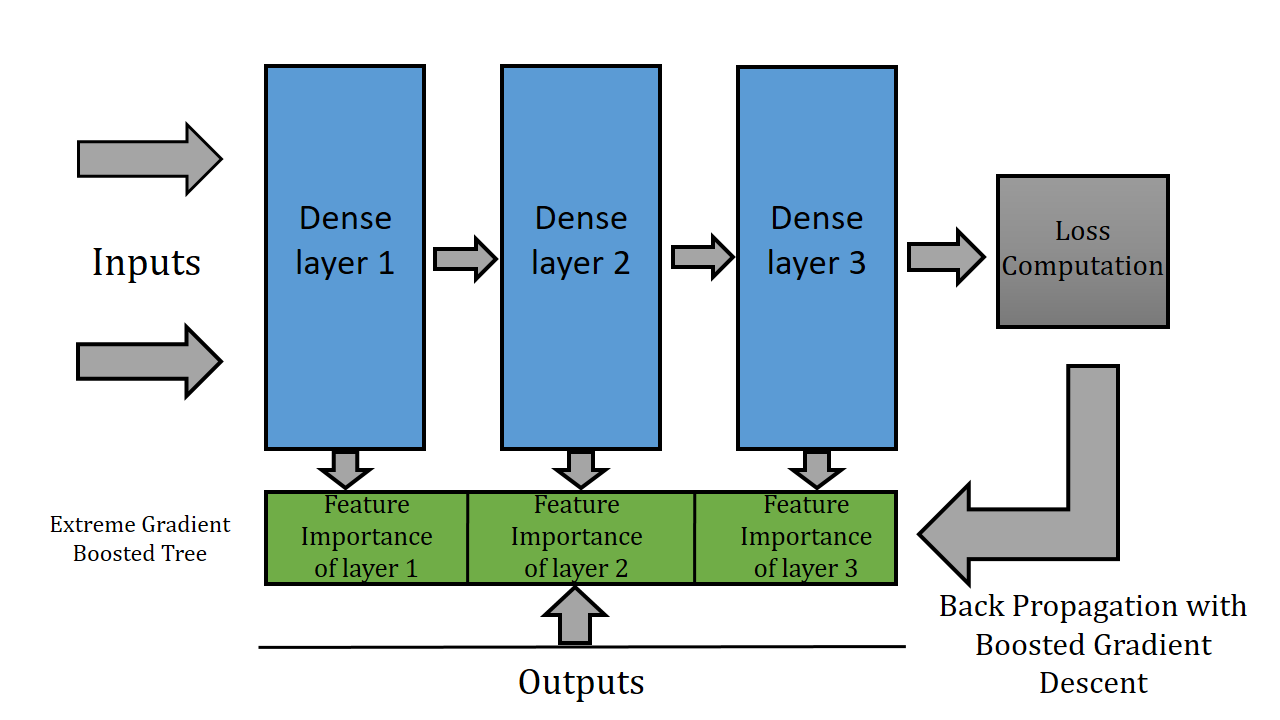}}
\caption{XBNet architecture}    
  \label{fig:sample_image}
\end{figure*}

\section{XBNet Architecture}

\subsection{Feature importance from XGBoost}
XGBoost is an ensemble technique that uses \( \mathcal{N} \) trees for providing a prediction in the following way:
\begin{flalign}
y &= \psi(x)= \sum_{n=1}^{N}g_n(x)&&
\end{flalign}
where \(x\) and \(y\) denote the input features and outputs respectively. \(g_n(x)\) represents the score of leaf of the \(N^{th}\) tree. Also,  $g_n(x) \in M$, where \(M\) represents the set of all the scores. We then regularize to avoid overfitting:
\begin{flalign}
\mathcal{L}(\psi) &= \sum_{i}l(\hat{y^i},y^i) + \sum_{n}\delta(g_n)&&
\end{flalign}
where $l$ denotes the loss function, and $\delta(g_n)$ is defined as:
\begin{flalign}
\delta(g) &= \gamma T + \frac{1}{2}\lambda\sum_{i=1}^{T}w_i^2&&
\end{flalign}
 where $\lambda$ and $\gamma$ help in reducing overfitting by directing regularization. \(T\) and  \(w\) represent the number of leaves and their weights respectively.

The feature importance of the extremely gradient boosted tree which plays a great role in our architecture as well as training, is determined based on information gain from the features of the tree, which is a way of determining which attribute in a given set of feature vectors is most useful for distinguishing between the classes to be learned which in turn is dependent on entropy that is a common way of measuring impurity \cite{raileanu2004theoretical} \cite{li2006spatial}. Impurity measures the homogeneity of the target variable at every node.
Let \( \mathcal{P} \) be a probability distribution such that
\begin{flalign}
P &= (p_1,p_2,....,p_n)&&
\end{flalign}
where \(p_i\) is the probability of a data point that belongs to a subset \(d_i\) of dataset \(\mathcal{D}\)
Entropy can be defined as:
\begin{flalign}
Entropy(P) &= \sum_{i=1}^{n}-p_i\log_2(p_i)&&
\end{flalign}
The information gain calculated is then used to determine the feature importance of the boosted tree which is used in Boosted Gradient Descent.
\begin{flalign}
Information Gain &= Entropy_{beforeSplit} - Entropy_{afterSplit}&&
\end{flalign}

\subsection{Gradient Descent}
Gradient Descent is an optimization technique for finding the minima of the required differentiable loss function. Gradient descent is employed to determine the optimum values of a function's parameters that minimize the given loss function. Forward propagation happens in the way shown below:
\begin{flalign}
z^{[l]} &= w^{[l]}A^{[l-1]} + b^{[l]}&&\\
A^{[l]} &= g^{[l]}(z^{[l]})&&
\end{flalign}
where $w^{[l]}$ and $b^{[l]}$ is the weight and the bias of the $l^{th}$ layer respectively $g(x)$ is the activation function, $A^{[l]}$ is the output of the $l^{th}$ layer that is activated with the activation function $g(x)$ and $z^{[l]}$ is the output of the $l^{th}$ layer before applying the activation function.Then we compute the cost 
\begin{flalign}
J &= \frac{1}{m}\sum\mathcal{L}(\hat{y}^{(i)},y^{(i)}) + \frac{\lambda}{2m}\sum (||w^{[l]}||)_f^{2}&&
\end{flalign}
where $\hat{y}^{(i)},y^{(i)}$ are the predicted and actual values respectively, m is number of mini-batches of data during training and $\lambda$ is the tuning parameter that controls the effect of the regularization. The presence of a regularization reduces overfitting by regulating penalty. Here we have used the $L2$ regularization that is proportional to $(||(w^{[l]}||)^{2}$. Weights are adjusted during the backward propagation in the following way:
\begin{flalign}
w^{[l]} &= w^{[l]} - \alpha\nabla w^{[l]}&&\\
b^{[l]} &= b^{[l]} - \alpha\nabla b^{[l]}&&
\end{flalign}
where $w^{[l]}$ and $b^{[l]}$ is the weight and the bias of the $l^{th}$ layer respectively, $\alpha$ is the learning rate and $\nabla w$ represents the gradient.

Given below is the algorithm for gradient descent where we combine all the steps to provide the flow of this optimization technique.
\begin{algorithm}[h!]
\SetAlgoLined
\KwResult{  Cost function is minimized using Gradient Descent }
 Initialize w, $\alpha$\;
 \For{t= 1,2,..,m}{
    Forward propagation on $X^{t}$\;{
             \hspace{5mm} $z^{[l]} = w^{[l]}A^{[l-1]} + b^{[l]} $ \;
             \hspace{5mm} $A^{[l]} = g^{[l]}(z^{[l]})$\;
    Compute cost J= \[\frac{1}{m}\sum\mathcal{L}(\hat{y}^{(i)},y^{(i)}) + \frac{\lambda}{2m}\sum (||w^{[l]}||)_f^{2}\]\
    Backward propagation on $J^{t}$\;
            \hspace{5mm} $w^{[l]} = w^{[l]} - \alpha\nabla w^{[l]} $ \;
            \hspace{5mm} $b^{[l]} = b^{[l]} - \alpha\nabla b^{[l]} $ \;
    }
 }
 
 \caption{Basic Gradient Descent approach}
\end{algorithm}
Our optimization strategy will add another step to the traditional gradient descent approach to boost the performance.

\subsection{Training and Optimization using Boosted Gradient Descent }
Now we combine the above subsections to create our architecture and optimizer. This optimizer can be used with any gradient-based optimization technique as the base but here we have given the example of Gradient Descent for easily elucidating the concept. Our architecture creates a Sequential structure of layers with the first and last being the input and output layers respectively. The weights of the first layer are not initialized randomly but it is the feature importance of a gradient boosted tree which is trained at the time of initialization of the model. 
\begin{flalign}
w^{[1]} &= tree.train(X,y).importance &&
\end{flalign}
Here $w^{1}$ is the weight of the first layer, X and y are the input and output vectors, respectively. Apart from this the architecture also contains a gradient boosted tree that is connected to each layer. At the time of training the model, the data that is fed completes a forward and backward propagation, and the weights of all the layers get updated according to gradient descent once and then instead of going to the next epoch of training it goes through all the layers again and updates its weights again based on the feature importance of the gradient boosted tree that is trained on the layers respectively. This is the step that is added during the forward propagation:
\begin{flalign}
f^{[l]} &= tree.train(A^{[l]},y^{(i)}).importance&&
\end{flalign}
where $A^{[l]}$ is the output of the $l^{th}$ layer that is activated with an activation function $g(x)$, $y^{(i)}$ is the outputs for the mini-batch that is fed to the system.
The number of layers of the neural network on which the tree should be trained is a hyperparameter. Further, the trees are trained on the hidden layers during the forward propagation of the feed-forward neural network and their feature importance is stored which is updated after the backward pass. So in this approach, the feature importance also plays the role of adjusting the weight which boosts the performance of the architecture. Weights are updated in the following way: 
\begin{flalign}
w^{[l]} &= w^{[l]} - \alpha\nabla w^{[l]}&&
\end{flalign}
where $w^{[l]}$ and $b^{[l]}$ is the weight and the bias of the $l^{th}$ layer respectively, $\alpha$ is the learning rate and $\nabla w$ represents the gradient.
\begin{flalign}
w^{[l]} &= w^{[l]} + f^{[l]}\phi(w^{[l]})&&
\end{flalign}
where $f^{[l]}$ is the feature importance for the $l^{th}$ layer that was computed during the forward propagation.
\begin{flalign}
\phi(w^{[l]}) &= 10^{log(\min (w^{[l]}))}&&
\end{flalign}
Here $\phi(w^{[l]})$ is used to ensure that the contribution of the weights provided by the feature importance and the weights of gradient descent is in the same order. The feature importance is scaled down to the same power as that of the minimum value of the weights of the gradient descent algorithm to ensure that the feature importance complements the weights obtained through gradient descent. This is necessary because after some epochs the feature importance remains in the same order by virtue of its definition but the weights provided by gradient descent decrease by several orders. Only one gradient boosted tree is initialized inside the architecture as all the layers will have different inputs after the epoch and therefore the same tree is used for each layer and each epoch and the feature importance are stored before training the tree on the next layer which saves space while providing the required result. In the algorithm given below, $\hat{y}^{(i)},y^{(i)}$ are the predicted and actual values respectively.
\begin{algorithm}[hbt!]
\SetAlgoLined
\KwResult{Cost function is minimized using Boosted Gradient Descent }
 Initialize w,b, $\alpha$,tree\;
 $w^{[1]} = tree.train(X,y$).importance\;
 \For{t= 1,2,..,m}{
    Forward propagation on $X^{t}$\;{
             \hspace{5mm} $z^{[l]} = w^{[l]}A^{[l-1]} + b^{[l]} $ \;
             \hspace{5mm} $A^{[l]} = g^{[l]}(z^{[l]})$\;
             \hspace{5mm} $f^{[l]} = tree.train(A^{[l]},y^{(i)}$).importance\;
    Compute cost J= \[\frac{1}{m}\sum\mathcal{L}(\hat{y}^{(i)},y^{(i)}) + \frac{\lambda}{2m}\sum (||w^{[l]}||)_f^{2}\]\
    Backward propagation on $J^{t}$\;
            \hspace{5mm} $w^{[l]} = w^{[l]} - \alpha\nabla w^{[l]} $ \;
            \hspace{5mm} $f^{[l]} = f^{[l]}\times 10^{log(\min (w^{[l]}))}$\;
            \hspace{5mm} $w^{[l]} = w^{[l]} + f^{[l]}$\;
            \hspace{5mm} $b^{[l]} = b^{[l]} - \alpha\nabla b^{[l]} $ \;
            
    }
 }
 
 \caption{Training Algorithm for XBNet using Boosted Gradient Descent}
\end{algorithm}

Thus each layer has two components, the weight vector and the matrix containing the feature importances. The configuration of each layer is such that during training of a particular layer the weights are first calculated during forward propagation, which is followed by training of an XGBoost model taking inputs as the nodes of that layer, the feature importance of this ensemble model is stored in a matrix as they are required again during back-propagation. The weights are updated on the basis of gradient descent during back-propagation then they are adjusted according to the matrix of feature importances that was calculated during forward propagation. The results section contains a detailed discussion of the different configurations that were tried as well as their performance. 

\subsection{Inference}
During inference, there is no requirement of using the gradient boosted tree as the purpose of the feature importance of the tree is to only ensure the precise update of weights to improve the performance of the network which is done while training the model. Hence the prediction speed only depends on the number of layers as there is no contribution of trees during inference. 

\subsection{Complexity}
 To understand the complexity of our proposed architecture, we will delve into the details of the complexities of the gradient boosted tree and the neural network.
 The training time and space complexity of a gradient boosted tree can be defined in the following way:
 \begin{flalign}
 Complexity_{time} = O(tn_{trees}log(td)) \\
 Complexity_{space} = O(n_{nodes}n_{trees} + \beta m)
 \end{flalign}
 
 where $t$, $n_{trees}$ and $d$ represent the number of data points, trees, and dimensions respectively; $n_{nodes}$ represents the number of nodes, and $\beta m$ represents the values that are processed by the leaves in the decision trees. As we are using the gradient boosted tree only for training the neural network, the complexities of the tree during testing are not mentioned.
 The training complexity of a deep neural network can be defined as follows:
 
 \begin{flalign}
 Complexity_{time} = O(nt(\sum_{i=1}^{p-1}\phi ({l}^{[i]}) \phi ({l}^{[i+1]}))) \\
 Complexity_{space} = O(2q + 1)
 \end{flalign}
 The testing complexity of a deep neural network can be defined as follows:
 
 \begin{flalign}
 Complexity_{time} = O(t(\sum_{i=1}^{p-1}\phi ({l}^{[i]}) \phi ({l}^{[i+1]}))) \\
 Complexity_{space} = O(2q + 1)
 \end{flalign}
 
  where $n$, $t$ represent the number of epochs and training pairs, respectively; $l^{[i]}$ represents the layer that is being processed, $\phi (x)$ number of nodes in the $x$th layer, $q$ represents the vector containing all the parameters and $p$ is the number of layers in the neural network.
  Now we derive the complexity of our architecture in the following way:
  \begin{enumerate}
      \item[(1)] During Training:
      \begin{flalign}
      Complexity_{time} = O(nt[(\sum_{i=1}^{p-1}\phi ({l}^{[i]}) \phi ({l}^{[i+1]})) + n_{trees}plog(td)]) \\
      Complexity_{space} = O(3q + n_{nodes}n_{trees} + \beta m + 1)
      \end{flalign}
      For every layer, we train a gradient boosted tree, so the time complexity of each layer is added by that factor multiplied by the number of layers as indicated in the above notation. For space complexity, the complexities of the gradient boosted tree and neural network are added, along with the matrix for storing the feature importances during the training procedure.
        \item[(2)] During Inference:
        \begin{flalign}
      Complexity_{time} = O(t[\sum_{i=1}^{p-1}\phi ({l}^{[i]}) \phi ({l}^{[i+1]}))) \\
      Complexity_{space} = O(2q + 1)
      \end{flalign}
      The space complexity decreases as the gradient boosted tree is only used while training and thus can be discarded during inference. The time complexity during inference is the same as a vanilla neural network.
  \end{enumerate}
  where $n$, $t$ represent the number of epochs and training pairs, respectively; $l^{[i]}$ represents the layer that is being processed, $\phi (x)$ number of nodes in the $x$th layer, $q$ represents the vector containing all the parameters; $p$ is the number of layers in the neural network, $d$ denotes the number of dimensions, $n_{nodes}$ represents the number of nodes, and $\beta m$ represents the values that are processed by the leaves in the decision trees and $n_{trees}$ represents the number of trees.

\section{Results}
\begin{figure*}[bt!]
\centerline{\includegraphics[width=11cm]{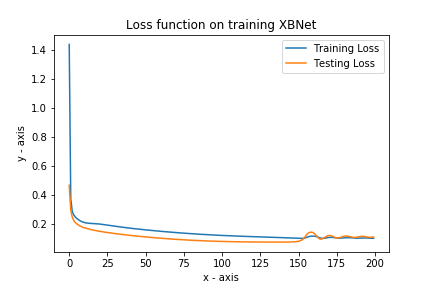}}
\caption{BCE Loss function vs Number of epochs}    
  \label{fig:loss}
\end{figure*}

\begin{figure*}[ht!]
\centerline{\includegraphics[width=11cm]{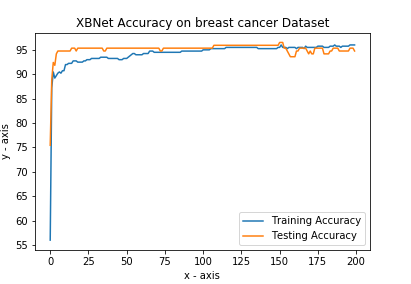}}
\caption{Accuracy vs Number of epochs}    
  \label{fig:acc}
\end{figure*}

\begin{figure*}[ht!]
\centerline{\includegraphics[width=14cm, height= 9cm]{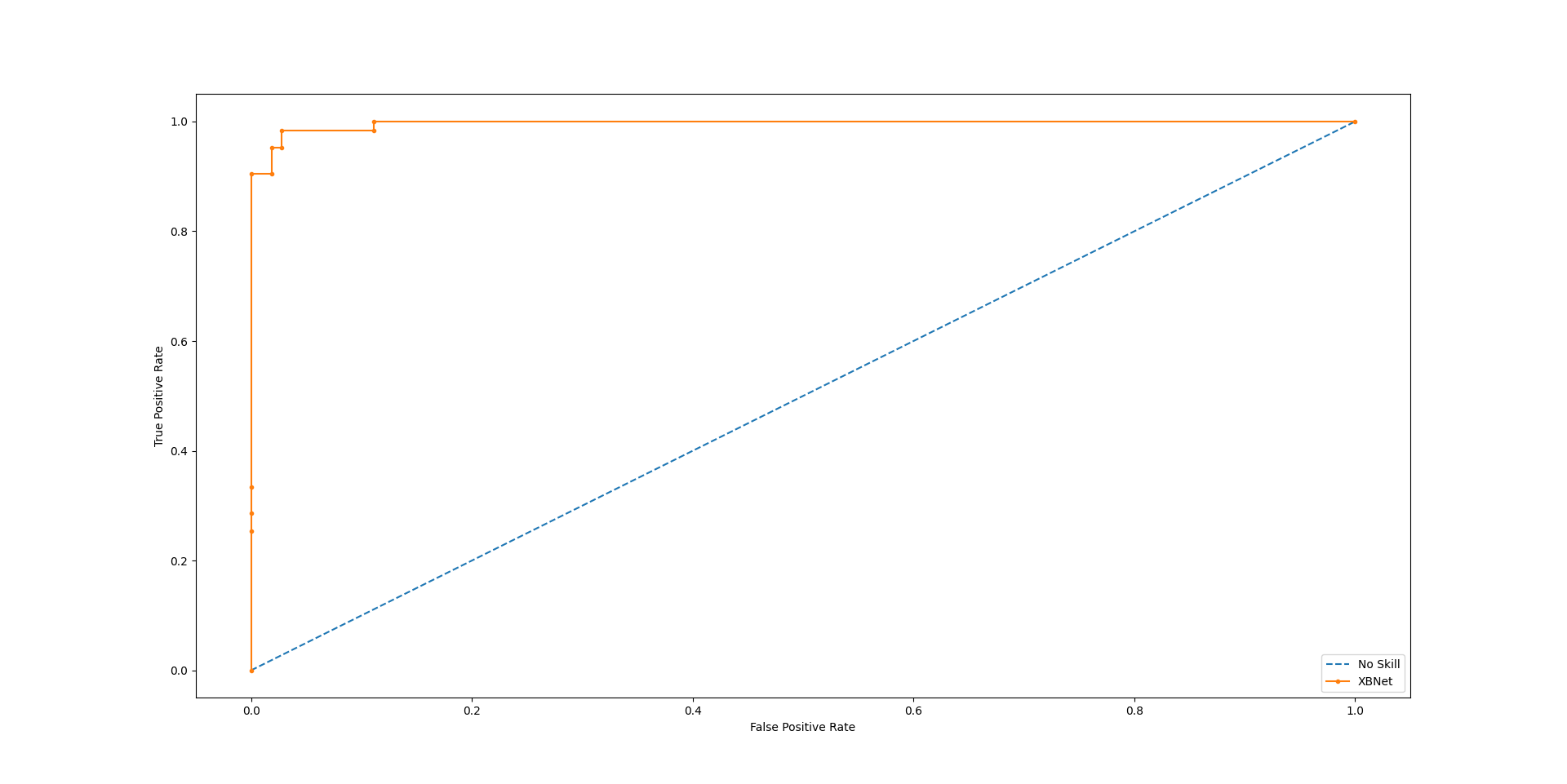}}
\caption{ROC of XBNet}    
  \label{fig:acc}
\end{figure*}

\begin{figure*}[ht!]
\centerline{\includegraphics[width=14cm, height= 9cm]{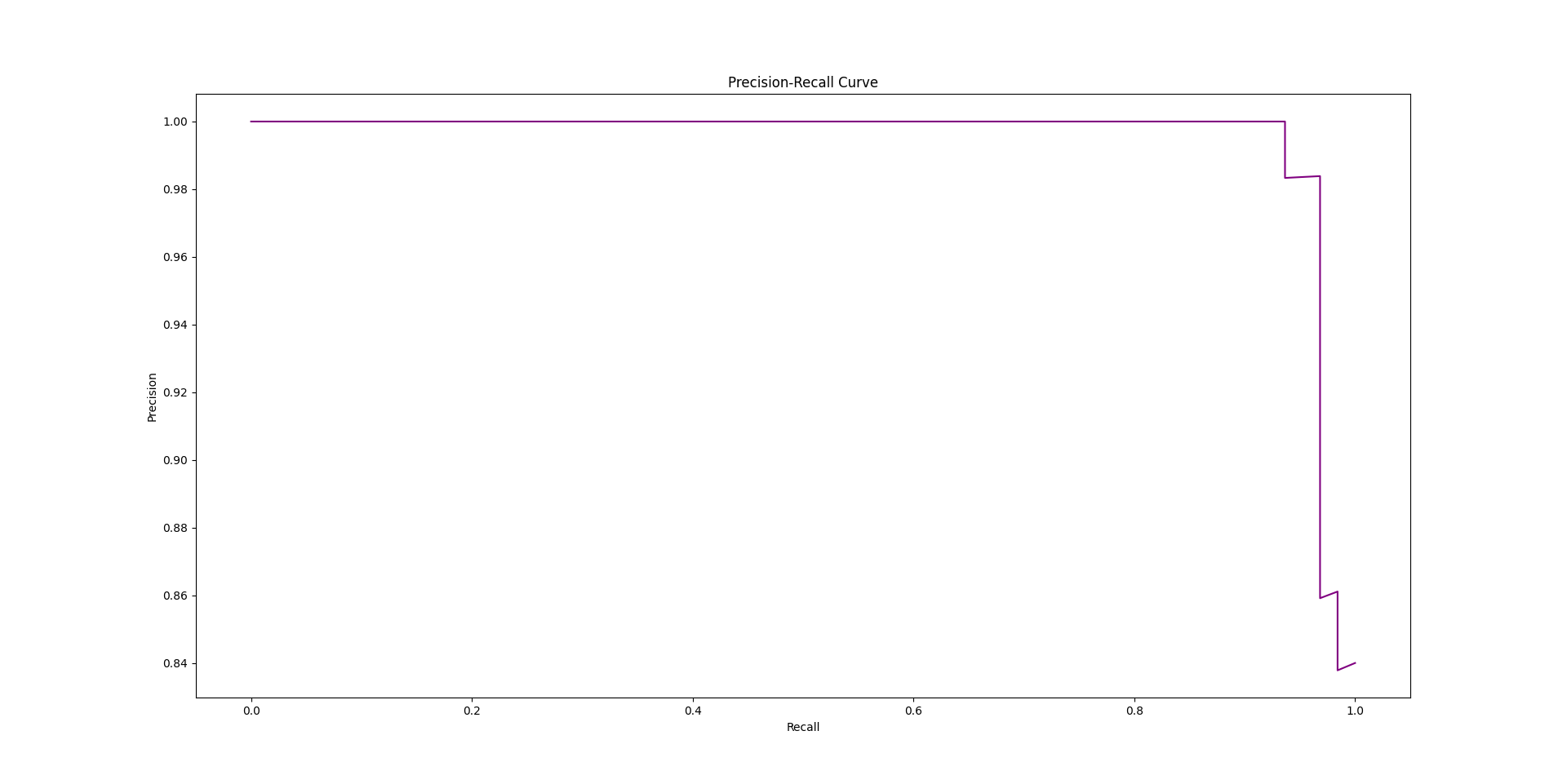}}
\caption{Precision Recall curve of XBNet}    
  \label{fig:acc}
\end{figure*}
We evaluate our model on the Breast Cancer Dataset and we create several models with different numbers of layers, numbers of nodes, activation functions, and different numbers of boosted layers i.e number of hidden layers on which the tree is trained. Here is a summary of the results:
\begin{enumerate}
\item[(1)] When a model with 2 layers having 16,1 nodes respectively whose 1st layer is boosted with a default xgboost tree with no hyperparameter tuning is trained on 100 epochs it yields a loss of 0.12 on the training data and a loss of 0.08 on the validation data when BCE loss criterion was used as the loss function.
\item[(2)] When a model with 2 layers having 8,1 nodes respectively whose 1st layer is boosted with a default xgboost tree with no hyperparameter tuning is trained on 100 epochs it yields a loss of 0.10 on the training data and a loss of 0.09 on the validation data when BCE loss criterion was used as the loss function.
\item[(3)] When a model with 2 layers having 8,1 nodes respectively whose 1st as well as 2nd layer is boosted with a default xgboost tree with no hyperparameter tuning is trained on 100 epochs it yields a loss of 0.13 on the training data and a loss of 0.11 on the validation data when BCE loss criterion was used as the loss function.
\item[(4)] When a model with 3 layers having 32,16,1 nodes respectively whose 1st layer is boosted with a default xgboost tree with no hyperparameter tuning is trained on 100 epochs it yields a loss of 21.78 on the training data and a loss of 19.45 on the validation data when BCE loss criterion was used as the loss function.\\

\end{enumerate}

ROC AUC score of 99.6\% was obtained on the breast cancer dataset.
The optimal settings that we established were 100 estimators in the gradient boosted tree, ReLU activation after each layer, a fewer number of layers, and fewer nodes in each layer in comparison to a vanilla neural network and boosting applied to the first few layers. We also introduced a parameter $\epsilon$ and set its value to 0.001 to introduce Laplacian smoothing in the feature importances. Adam optimizer with a learning rate of 0.01 is used and the train-test split was stratified with 80\% of the data being used for training and the rest for testing; $mlogloss$ which computes log loss for multiclass problems was used as the evaluation metric for the gradient boosted trees.
\\


\begin{table}%
 \caption{Results on different Datasets}
\begin{tabularx}{0.8\textwidth} { 
  | >{\centering\arraybackslash}X 
  | >{\centering\arraybackslash}X 
  | >{\centering\arraybackslash}X 
  | >{\centering\arraybackslash}X | }
 \hline
Dataset & Training Accuracy & Testing Accuracy    \\
 \hline
 Iris & 100 &  100  \\
  \hline
 Breast Cancer & 96.7 &  96.49  \\
 \hline
  Wine  & 97.22 &  97.22  \\
 \hline
 Diabetes & 77.09 &  78.78  \\
 \hline
 Titanic & 80.25 &  79.85  \\
 \hline
 Digits & 99.65 &  94.72  \\
 \hline
  German Credit & 69.8 &  71.33  \\
 \hline
  Digit Completion & 86.11 &  85.98  \\
 \hline
 
 \end{tabularx}

 \end{table}%
 
 \begin{table}%
 \caption{Training Classification metrics on Breast Cancer}
\begin{tabularx}{0.8\textwidth} { 
  | >{\centering\arraybackslash}X 
  | >{\centering\arraybackslash}X 
  | >{\centering\arraybackslash}X 
  | >{\centering\arraybackslash}X | }
 \hline
 Class & Precision & Recall & f1-score \\
 \hline
 0 & 0.96 &  0.98 & 0.97  \\
 \hline
 1 & 0.97 &  0.93 & 0.95   \\
 \hline
 micro avg & 0.96 &  0.96 & 0.96   \\
 \hline
 macro avg & 0.96 &  0.96 & 0.96   \\
 \hline
 weighted avg & 0.96 &  0.96 & 0.96  \\
 \hline

\end{tabularx}
\end{table}%

 \begin{table}%
 \caption{Testing Classification metrics on Breast Cancer}
\begin{tabularx}{0.8\textwidth} { 
  | >{\centering\arraybackslash}X 
  | >{\centering\arraybackslash}X 
  | >{\centering\arraybackslash}X 
  | >{\centering\arraybackslash}X | }
 \hline
 Class & Precision & Recall & f1-score \\
 \hline
 0 & 0.99 &  0.93 & 0.96  \\
 \hline
 1 & 0.89 &  0.98 & 0.93   \\
 \hline
 micro avg & 0.95 &  0.95 & 0.95   \\
 \hline
 macro avg & 0.94 &  0.96 & 0.94   \\
 \hline
 weighted avg & 0.95 &  0.95 & 0.95   \\
 \hline

\end{tabularx}

\end{table}%

\begin{table}%
 \caption{Performance Comparision of XBNet and XGBoost}
\begin{tabularx}{0.8\textwidth} { 
  | >{\centering\arraybackslash}X 
  | >{\centering\arraybackslash}X 
  | >{\centering\arraybackslash}X 
  | >{\centering\arraybackslash}X | }
 \hline
Dataset & XBNet & XGBoost    \\
 \hline
 Iris & 100 &  97.7  \\
  \hline
 Breast Cancer & 96.49 &  96.47  \\
 \hline
  Wine  & 97.22 &  97.22  \\
 \hline
 Diabetes & 78.78 &  77.48  \\
 \hline
 Titanic & 79.85 &  80.5  \\
 \hline
  German Credit & 71.33 &  77.66  \\
 \hline
  Digit Completion &  85.98 & 78.24 \\
 \hline
 
 \end{tabularx}

 \end{table}%

\section{Limitations and Future Work}
XBNet requires more time and resources for training as we train a gradient boosted tree in every layer. We minimize the training time of this gradient boosted tree by fixing the number of trees as 100. The optimal number of trees for training in each layer can also be parameterized to find the best-suited number to minimize the number of trees and concurrently improve the evaluation. Currently, XBNet only works on tabular data and is unable to process unstructured data. If it is extended for usage in unstructured data, the number of parameters will see a drastic jump, and hence care has to be taken with respect to it.

\section{Conclusion}
This paper about XBNet discussed the techniques that we employed for building this architecture and described the training, optimization, and inference techniques for this model. As the need and use for data keep increasing day by day in our current world and its impact on the daily life of people is motivating data-driven decisions in many sectors of the industry, this paper was an effort to combine neural networks and gradient boosted trees to provide an alternative approach to the currently used techniques which will pave the way for future work using this approach. The performance, interpretability, and scalability of this architecture will make it possible for professionals to optimally utilize the model.

\section{Acknowledgements}
I would like to thank Chandan Sarkar, Mallika Sarkar, Disha Shah, Vaibhav Vasani, and Dr.Rupali Patil for their constant guidance and valuable feedback. I am also grateful to Aparna Sarkar, Sneha Kothi, and the entire XBNet community for their priceless suggestions which went a long way toward improving the architecture. 


\bibliography{mybibfile}

\begin{thebibliography}{10}
\expandafter\ifx\csname url\endcsname\relax
  \def\url#1{\texttt{#1}}\fi
\expandafter\ifx\csname urlprefix\endcsname\relax\def\urlprefix{URL }\fi
\expandafter\ifx\csname href\endcsname\relax
  \def\href#1#2{#2} \def\path#1{#1}\fi

\bibitem{jordan2015machine}
M.~I. Jordan, T.~M. Mitchell, Machine learning: Trends, perspectives, and
  prospects, Science 349~(6245) (2015) 255--260.

\bibitem{wang2010don}
A.~H. Wang, Don't follow me: Spam detection in twitter, in: 2010 international
  conference on security and cryptography (SECRYPT), IEEE, 2010, pp. 1--10.

\bibitem{friedman2001greedy}
J.~H. Friedman, Greedy function approximation: a gradient boosting machine,
  Annals of statistics (2001) 1189--1232.

\bibitem{he2015delving}
K.~He, X.~Zhang, S.~Ren, J.~Sun, Delving deep into rectifiers: Surpassing
  human-level performance on imagenet classification, in: Proceedings of the
  IEEE international conference on computer vision, 2015, pp. 1026--1034.

\bibitem{devlin2018bert}
J.~Devlin, M.-W. Chang, K.~Lee, K.~Toutanova, Bert: Pre-training of deep
  bidirectional transformers for language understanding, arXiv preprint
  arXiv:1810.04805.

\bibitem{lecun2015deep}
Y.~LeCun, Y.~Bengio, G.~Hinton, Deep learning. nature 521 (7553), 436-444,
  Google Scholar Google Scholar Cross Ref Cross Ref.

\bibitem{mobahi2009deep}
H.~Mobahi, R.~Collobert, J.~Weston, Deep learning from temporal coherence in
  video, in: Proceedings of the 26th Annual International Conference on Machine
  Learning, 2009, pp. 737--744.

\bibitem{chui2018notes}
M.~Chui, J.~Manyika, M.~Miremadi, N.~Henke, R.~Chung, P.~Nel, S.~Malhotra,
  Notes from the ai frontier: Insights from hundreds of use cases, McKinsey
  Global Institute.

\bibitem{hooker2018evaluating}
S.~Hooker, D.~Erhan, P.-J. Kindermans, B.~Kim, Evaluating feature importance
  estimates.

\bibitem{hastie2009multi}
T.~Hastie, S.~Rosset, J.~Zhu, H.~Zou, Multi-class adaboost, Statistics and its
  Interface 2~(3) (2009) 349--360.

\bibitem{malioutov2017learning}
D.~M. Malioutov, K.~R. Varshney, A.~Emad, S.~Dash, Learning interpretable
  classification rules with boolean compressed sensing, in: Transparent Data
  Mining for Big and Small Data, Springer, 2017, pp. 95--121.

\bibitem{pedregosa2011scikit}
F.~Pedregosa, G.~Varoquaux, A.~Gramfort, V.~Michel, B.~Thirion, O.~Grisel,
  M.~Blondel, P.~Prettenhofer, R.~Weiss, V.~Dubourg, et~al., Scikit-learn:
  Machine learning in python, the Journal of machine Learning research 12
  (2011) 2825--2830.

\bibitem{breiman2001random}
L.~Breiman, Random forests, Machine learning 45~(1) (2001) 5--32.

\bibitem{10.1145/2939672.2939785}
T.~Chen, C.~Guestrin, \href{https://doi.org/10.1145/2939672.2939785}{Xgboost: A
  scalable tree boosting system}, in: Proceedings of the 22nd ACM SIGKDD
  International Conference on Knowledge Discovery and Data Mining, KDD '16,
  Association for Computing Machinery, New York, NY, USA, 2016, p. 785–794.
\newblock \href {http://dx.doi.org/10.1145/2939672.2939785}
  {\path{doi:10.1145/2939672.2939785}}.
\newline\urlprefix\url{https://doi.org/10.1145/2939672.2939785}

\bibitem{rota2014neural}
S.~Rota~Bulo, P.~Kontschieder, Neural decision forests for semantic image
  labelling, in: Proceedings of the IEEE Conference on Computer Vision and
  Pattern Recognition, 2014, pp. 81--88.

\bibitem{yang2018deep}
Y.~Yang, I.~G. Morillo, T.~M. Hospedales, Deep neural decision trees, arXiv
  preprint arXiv:1806.06988.

\bibitem{ruder2016overview}
S.~Ruder, An overview of gradient descent optimization algorithms, arXiv
  preprint arXiv:1609.04747.

\bibitem{thongsuwan2021convxgb}
S.~Thongsuwan, S.~Jaiyen, A.~Padcharoen, P.~Agarwal, Convxgb: A new deep
  learning model for classification problems based on cnn and xgboost, Nuclear
  Engineering and Technology 53~(2) (2021) 522--531.

\bibitem{sugiharti2021convolutional}
E.~Sugiharti, R.~Arifudin, D.~Wiyanti, A.~Susilo, Convolutional neural
  network-xgboost for accuracy enhancement of breast cancer detection, in:
  Journal of Physics: Conference Series, Vol. 1918, IOP Publishing, 2021, p.
  042016.

\bibitem{geng2016learning}
Y.~Geng, R.-Z. Liang, W.~Li, J.~Wang, G.~Liang, C.~Xu, J.-Y. Wang, Learning
  convolutional neural network to maximize pos@ top performance measure, arXiv
  preprint arXiv:1609.08417.

\bibitem{raileanu2004theoretical}
L.~E. Raileanu, K.~Stoffel, Theoretical comparison between the gini index and
  information gain criteria, Annals of Mathematics and Artificial Intelligence
  41~(1) (2004) 77--93.

\bibitem{li2006spatial}
X.~Li, C.~Claramunt, A spatial entropy-based decision tree for classification
  of geographical information, Transactions in GIS 10~(3) (2006) 451--467.

\end{thebibliography}

\end{document}